\documentclass{article}

\usepackage[preprint]{neurips_2023}

\usepackage[utf8]{inputenc} 
\usepackage[T1]{fontenc}    
\usepackage{hyperref}       
\usepackage{url}            
\usepackage{booktabs}       
\usepackage{amsfonts}       
\usepackage{nicefrac}       
\usepackage{microtype}      
\usepackage{xcolor}         

\usepackage[ruled,linesnumbered]{algorithm2e}
\usepackage{graphicx}
\usepackage{colortbl}
\usepackage{makecell}
\usepackage{booktabs}
\usepackage{multirow}
\usepackage{float}
\title{HiveNAS: Neural Architecture Search using Artificial Bee Colony Optimization}

\author{
  Mohamed Shahawy \\
  Department of Computer Science\\
  Staffordshire University\\
  Staffordshire, UK\\
  \texttt{mohamed.shahawy@research.staffs.ac.uk} \\
  \And
   Elhadj Benkhelifa \\
  Department of Computer Science\\
  Staffordshire University\\
  Staffordshire, UK\\
  \texttt{e.benkhelifa@staffs.ac.uk} \\
}

\def\BibTeX{{\rm B\kern-.05em{\sc i\kern-.025em b}\kern-.08em
    T\kern-.1667em\lower.7ex\hbox{E}\kern-.125emX}}
    
\makeatletter\let\MPtrue\@minipagetrue\makeatother
    
\newcolumntype{P}[1]{>{\centering\arraybackslash}p{#1}}
\newcolumntype{M}[1]{>{\centering\arraybackslash}m{#1}}

\begin{document}

\maketitle

\begin{abstract}
  The traditional Neural Network-development process requires substantial expert knowledge and relies heavily on intuition and trial-and-error. Neural Architecture Search (NAS) frameworks were introduced to robustly search for network topologies, as well as facilitate the automated development of Neural Networks. While some optimization approaches -- such as Genetic Algorithms -- have been extensively explored in the NAS context, other Metaheuristic Optimization algorithms have not yet been investigated. In this study, we evaluate the viability of Artificial Bee Colony optimization for Neural Architecture Search. Our proposed framework, HiveNAS, outperforms existing state-of-the-art Swarm Intelligence-based NAS frameworks in a fraction of the time.
\end{abstract}

\section{Introduction}

In the recent years, Deep Neural Networks (DNNs) have become integral in numerous domains due to their astonishing success and wide range of applications. The DNN design process, however, requires several high-level decisions (network depth, layer type, hyperparameters, connectivity, etc) which translate into search spaces typically containing hundreds of thousands/millions of possible solutions. Attempting to manually find an optimal solution in such a massive space is becoming increasingly challenging; humans tend to have a fixed and biased thinking paradigm, which leads to (i) unvaried design patterns, (ii) a time-consuming development process (based primarily on trial-and-error/intuition), and (iii) dependence on domain-specific knowledge.

Neural Architecture Search (NAS), a subfield of AutoML, emerged decades ago to address the drawbacks posed by the manual, human-dependent architecture design process \cite{miller_1989}. It was not until recently, however, that NAS successfully outperformed humans \cite{zoph_2017,zoph_2018}. NAS leverages existing optimization algorithms to traverse the enormous architectures' solutions space and robustly converge on optimal topologies in an automated, domain-agnostic fashion. While Hyperparameter Optimization is considered a field on its own, it largely overlaps with NAS as both objectives can be optimized simultaneously \cite{feurer_2019}.

A NAS framework is generally composed of 3 main components: a \textit{Search Space}, a \textit{Search Algorithm}, and an \textit{Evaluation Strategy} \cite{elsken_2019}. The Search Space defines all the possible candidate networks that can be sampled, while the Search Algorithm navigates said space to find an optimal architecture. The Evaluation Strategy guides the Search Algorithm by providing a fitness score for each sampled candidate (e.g. training a candidate network and returning its performance metrics).

\subsection{Motivation \& Contributions}

In this study, we introduce the first Neural Architecture Search framework based on Artificial Bee Colony (ABC) optimization, dubbed \textit{HiveNAS}. The ABC algorithm features a number of desirable properties for NAS models, including: efficient parallel exploration of the solution space, dynamic exploitation rate (probabilistic convergence based on fitness scores), and scalable/distributed computing capabilities.

Our experiments show that ABC is an effective search algorithm for Neural Architecture Search frameworks, can design architectures with comparable results to the current state-of-the-art, and out-performs existing Swarm Intelligence (SI) NAS frameworks. In addition to investigating the viability of Artificial Bee Colony optimization for NAS, several secondary contributions are also introduced, including:

\begin{enumerate}
    \item Open-sourcing HiveNAS\footnote{https://github.com/ThunderStruct/HiveNAS/} for further investigative research opportunities
    \item Thoroughly documenting the framework\footnote{https://hivenas.readthedocs.io/en/latest/} for potential real-world usage and adoption
    \item Conducting experiments using several different configuration setups to empirically study the effects of NAS components (network depth, operation types, epochs per candidate, etc.)
    \item Democratizing NAS through a user-friendly Google Colab interface\footnote{https://colab.research.google.com/github/ThunderStruct/HiveNAS/blob/main/colab/HiveNas.ipynb} to optimize neural architectures with minimal programming knowledge
\end{enumerate}

Furthermore, the framework is extensively configurable and is built using a modular design pattern (expanded upon in \ref{subsubsec:framework_design}) to facilitate any additional non-parameterized requirements.

In the next section, works from neighboring areas are briefly reviewed. Following which, the details of ABC optimization, the HiveNAS framework, and the experiments conducted are introduced. In Section \ref{sec:results}, the attained results are presented and discussed in comparison with the state-of-the-art. Finally, we conclude and delineate future directions for ABC-based NAS frameworks as well as general NAS components' potential improvements.

\section{Related Works}

While most NAS researches are predominantly based on Evolutionary Algorithms (EAs) and Reinforcement Learning (RL), SI algorithms have shown exceptional potential \cite{niu_2019,byla_2019} and several methods of which have not yet been explored \cite{shahawy_2022}. One such method is the stochastic, Metaheuristic Optimization (MO) algorithm called Artificial Bee Colony (ABC), which is inspired by the honeybees' foraging behavior (presented in Section \ref{subsec:abc}).

In general, MO algorithms (which include EAs and SI optimization) tend to have a number of favorable qualities for NAS problems, namely their parallel exploration capabilities and exceptional mutli-objective/high-dimensional optimization performance. Nevertheless, MO-based frameworks (non-evolutionary ones specifically) are often out-shined by RL in the literature. Simulated Annealing (SA), for instance, has outperformed most state-of-the-art NAS models \cite{mo_2021}, and yet has not been covered in most purportedly comprehensive surveys \cite{shahawy_2022}. Results from other MO-based NAS models, including Cuckoo Search and Tabu Search, also show promising potential and should be investigated further.

Ever since its inception in 2005, the Artificial Bee Colony optimization algorithm has been repeatedly successful in various domains \cite{karaboga_ABC_2007,karaboga_ABC_ANN_2007}. Similar to other SI algorithms, ABC uses a divide-and-conquer approach whereby a colony of bees represent different solutions (or candidates) in a given NP-hard search space. Although the classical ABC has yielded effective results on multiple high-dimensional numerical benchmarks, several variants have emerged that offer different control parameters (colony-size drop, adaptive abandonment limit, etc) and operators (selection, fitness evaluation, etc). We refer the reader to the recent, in-depth review by Sharma et al. \cite{sharma_2020}.

Artificial Bee Colony's fast convergence, flexibility, and wide range of exploration have been previously demonstrated in the \textit{ABCPruner} architecture compression framework \cite{lin_abcpruner_2021}. Although ABCPruner outperformed most state-of-the-art alternatives, Artificial Bee Colony is still eclipsed by the more mainstream optimization algorithms in the literature and its full potential is yet to be explored. Notwithstanding ABCPruner's effective capabilities, the framework is limited to compressing existing models rather than design architectures from scratch.

To further investigate the viability of ABC in the neural topology optimization domain, we propose the first Artificial Bee Colony-based Neural Architecture Search framework capable of designing and training ANNs with minimal human-intervention.

\section{Methodology}

In this section, we present our proposed framework, \textit{HiveNAS}. The general NAS foundation includes 3 phases: (1) designing a Search Space, (2) implementing a Search Algorithm that samples candidates from the given space, and (3) defining an Evaluation Strategy to estimate the performance of each candidate. Although the novelty of applying ABC over the high-dimensional neural architecture space is the main contribution proposed in this study, several secondary contributions are discussed in Section \ref{subsec:hivenas}.

\subsection{Artificial Bee Colony Optimization} \label{subsec:abc}

Artificial Bee Colony is a population-based stochastic optimization algorithm inspired by honeybees' foraging behavior, whereby a colony of bees searches for and converges on \textit{Food Sources} containing the most \textit{Nectar}. In the optimization context, Food Sources and Nectar represent positions on the optimization surface and their corresponding fitness scores, respectively.

To model the division of labor and self-organization that capture the essence of ABC, a bee colony is split into 3 types of bees, each having its own role:

\begin{enumerate}
    \item Scout Bees \textbf{(initialization)}: randomly search for food sources across the search space and initialize a set of optimization centers.
    \item Employee Bees \textbf{(exploration)}: Evaluate the initialized food sources or a neighboring position, then return to the hive and perform a “dance” (convey nectar information) to the colony.
    \item Onlooker Bees \textbf{(exploitation)}: Observe the employed bees’ dance and, following a fitness-based stochastic selection operator (e.g. roulette wheel), get assigned to a particular food source. The onlooker bees then search for and evaluate neighboring positions, creating a positive feedback loop around high-nectar areas
\end{enumerate}

\begin{algorithm}[ht]
\SetAlgoLined
\caption{Artificial Bee Colony Optimization for NAS}
\SetKw{Input}{input}
\SetKw{Initialization}{initialization}
\SetKw{Output}{Output}

\SetKw{EB}{Employee Bees}
\SetKw{OB}{Onlooker Bees}
\SetKw{GS}{Greedy Selection}
\SetKw{SB}{Scout Bees}

\Initialization{: Specify number of iterations T,
Generate colony,
Scout Bees initialize optimization centers,
Assign Employee Bees to initialized positions}
\BlankLine
\For{t = 1, 2, ..., T}{
    \EB{: 
    \BlankLine
    \For{employee in Employee Bees}{
        Evaluate Scout-sampled optimization center,
    \BlankLine
        Find neighboring position and evaluate it
    }}
    \BlankLine
    \OB{:     \BlankLine
    \For{onlooker in Onlooker Bees}{
        Using a stochastic selection operator (roulette wheel, for instance), assign onlooker to a high-fitness employee,
    \BlankLine
        Find neighboring position and evaluate it
    }}
    \BlankLine
    \GS{: (Elitism) Select and record highest performing candidate positions, 
    \BlankLine
    swap optimization centers with highest fitness neighbors
    \BlankLine
    \SB{:
    \For{scout in Scout Bees}{
        Check if linked employee bee has reached the Abandonment Limit,
    \BlankLine
        Reset employee accordingly to a newly sampled random position on the optimization surface
    }}
    }
    
}
\BlankLine
\Output{Highest performing candidate with its fitness scores}
\BlankLine
\end{algorithm}

\subsubsection{Initialization Phase}

The initialization of ABC starts with the Scouts randomly sampling a vector of starting food sources, $\vec{x}_{m}$, for $m=1,...,\mathcal{M}$, where $\mathcal{M}$ is the number of parallel explorations (scout bees' count).

Each food source, $\vec{x}_{m}$, represents a position on the optimization surface and is composed of $n$ components, $(x_{mi}\;; i=1, ... , n)$ (i.e $n$-dimensional position), which are used to control the optimization's convergence.

\begin{equation}
    x_{mi} = l_{i} + rand(0, 1) * (u_{i} - l_{i})
\label{eq:random_sampling}
\end{equation}

In the classical ABC algorithm, random initial sampling is modelled as a function of the upper and lower limits ($u_{i}$ and $l_{i}$, respectively) of each parameter of $x_{mi}$.

\subsubsection{Exploration Phase}

Once the scouts initialize the starting optimization centers, Employee Bees are then each assigned one food source. Although there is no explicit rule defining this 1-to-1 relationship between scouts/food-sources and employees, the convention dictates that Employee Bees are linked to one and only one position on the optimization surface \cite{sharma_2020}. A 1-to-many relationship between scouts and employees (or vice-versa) would introduce an optimization bias to particular positions over others, and while that may randomly coincide with better results, uniformity is generally a more favored approach.

Employee Bees are then tasked to randomly sample a neighboring position, typically modelled as:

\begin{equation}
    \upsilon_{mi} = x_{mi} + \phi_{mi}(x_{mi} - x_{ki})
\label{eq:neighbor_selection}
\end{equation}

Where $\upsilon_{mi}$ is the neighboring food source and $\phi_{mi}$ is a random factor used to offset a positional parameter.


Sampling operators in the classical ABC (Equations \ref{eq:random_sampling} \& \ref{eq:neighbor_selection}) are designed to be used on numerical benchmarks or other continuous spaces. They cannot, however, be used in the discrete NAS context without necessary alterations (see Section \ref{subsec:search_algorithm}).

\subsubsection{Exploitation Phase}

Finally, the Onlooker Bees' phase of ABC is a stochastic operation that favors elitism. By probabilistically favoring high-fitness positions, a positive feedback loop occurs around said positions whilst not completely abandoning low-fitness sources before realizing their potential.

\begin{equation}
    fit_m (\vec{x}_{m})= \left\{ {\begin{array}{*{20}c}
   {\frac{1}{{1 + f_m (\vec{x}_{m})}}}  & {} & {{\rm if}~~{\rm{ }}f_m(\vec{x}_{m})  \ge 0}  \\
   {1 + abs(f_m (\vec{x}_{m}))} & {}  & {{\rm if}~~{\rm{ }}f_m (\vec{x}_{m}) < 0}   \\
\end{array}} \right\}
\label{eq:fitness_evaluation}
\end{equation}

\begin{equation}
    p_m  = \frac{{fit_m(\vec{x_m}) }}{{\sum\limits_{m = 1}^{SN} {fit_m (\vec{x_m})} }}
    \label{eq:roulette_wheel}
\end{equation}

The classical ABC algorithm uses the \textit{fitness-proportionate} (also known as “roulette wheel”) selection operator to assign onlookers to employees. However, variants of the original ABC have used modified selection operators depending on the desired outcomes.

Each optimization center $\vec{x}_{m}$ is constrained by an “\textit{Abandonment Limit}, which specifies how many cumulative evaluations (by both employees and onlookers) are required before that area is considered “depleted” or sufficiently exploited. Once a food source is exhausted, the associated employee is then freed and a scout bee is tasked to sample a new random initial position, repeating the cycle for a given number of iterations.

\subsection{HiveNAS} \label{subsec:hivenas}

\subsubsection{Data Preprocessing}

To generalize the model and help narrow the gap between training and validation accuracies, HiveNAS incorporates several augmentation and regularization techniques in its pipeline.
A number of affine transformations are implemented in the framework, including:

\begin{enumerate}
    \item Rotation (up to $15^{\circ}$)
    \item Zoom/Crop (a $[0.8, 1.1]$ range factor, inclusive)
    \item Shear/Axis Distortion
    \item Height/width shift
    \item Horizontal flip
\end{enumerate}

In addition to affine transformations, HiveNAS can also be configured to apply (i) \textit{contrast}, (ii) \textit{saturation}, and (iii) \textit{cutout augmentations} \cite{devries_2017}. Each transformation and augmentation is applied image-wise and in real-time, with configurable probability, which significantly generalizes the model and improves its predictive capabilities. 

Manipulating each sample individually prior to passing it through the neural network, however, has a tremendous impact on computational resources and therefore on the entire architecture search process. By consuming more time per candidate evaluation, the framework’s convergence on optimal architectures is impeded. Nevertheless, these preprocessing techniques significantly improve the candidates' performance estimation and can therefore provide more realistic feedback to the optimization algorithm. The trade-off between accurate candidate measurement and overall performance is studied and discussed in Section \ref{sec:results}.

\subsubsection{Search Space} \label{subsubsec:search_space}

There are several NAS Search Space concepts in the literature that prioritize different qualities in the search process. Cell-based spaces, for instance, generally focus on micro-architecture search, where a manually defined structure of ``cells" is set and the Search Algorithm attempts to find the best performing operations for each cell type. Although Cell-based spaces tend to be less flexible, they tend to have an outstanding performance given adequate manual guidance. Layer-based spaces, on the other hand, are more exhaustive and therefore are less limited by predefined boundaries. Their tremendously larger solution space, however, requires more efficient exploration and exploitation.

To encode architectures and build the Search Space, most approaches generate a Directed Acyclic Graph (DAG) to capture all the possible solutions, which can be traversed by the Search Algorithm (as implemented in other SI-based NAS \cite{byla_2019}). For memory-efficiency purposes, HiveNAS discards the DAG-encoding concept and instead encodes architectures on-the-fly, storing each positional data (string-encoded neural architecture) in its corresponding bee memory. For instance, a single, isolated position on the optimization surface (a Food Source, $x_{m}$) holds all the necessary information required for further traversal in its local area (i.e. neighbor-sampling). In a typical experimental setup used in this study, whereby the Search Space comprised of $5$-layers deep networks, $10$ possible operations, and a colony size of $7$, the traditionally-encoded $100,000$-node DAG would be simply replaced by the $7$ currently evaluated candidates and a supplementary ``visited'' array. This approach yields identical traversal effectiveness of a DAG whilst preserving a substantial amount of memory for other operations.

Moreover, to enhance architectural diversity and address domain-specific requirements, HiveNAS implements both simple and compounded (where multiple layers are combined and treated as 1 operation) Layer-based Search Spaces, which can be configured as needed. Enabling hybrid simple and compounded layer-based architectures, however, results in a more expansive solution space, which is generally more difficult to optimize. Lastly, some (experimentally deduced) predefined compounded layers in the framework include residual blocks, which have yielded the top performing results for the models generated. Custom layer compounds, operations, and residual blocks can also be trivially added (delineated in the documentation) to fit program-specific requirements.


\subsubsection{Search Algorithm} \label{subsec:search_algorithm}

NAS Search Spaces are inherently discrete and therefore modified versions of the ABC sampling models (Equations \ref{eq:random_sampling} \& \ref{eq:neighbor_selection}) are required. Although some NAS framework relax the operations' discrete solution space into a continuous and differentiable optimization problem \cite{liu_2018}, it is more practical to augment the sampling operators in this case rather than convert the solution space.

Another complication that arises when applying ABC (as well as several other MO algorithms) to NAS is the definition of what a \textit{neighbor} is in the context of neural architectures. Neighboring individuals in MO should reflect a ``short distance" on the optimization surface and have reasonably similar fitness. Given 2 neural architectures, there is currently no way in the literature to have a priori estimation of their closeness on the solutions' surface. However, the generally acceptable assumption is that 2 architectures are neighbors if there is a 1-operation (i.e layer) difference between them \cite{white_2021}.

\subsubsection{Evaluation Strategy}

The primary deciding factor in a NAS framework's overall speed is its Evaluation Strategy, or how it estimates each candidate's performance (typically accuracy/error). The groundbreaking work by Zoph \& Le, for instance, fully trained each candidate architecture until convergence, which ended up taking over 500,000 GPU hours to find an optimal solution. Although full-training provides the most accurate estimation of a candidate's performance, the computational requirements for a such a framework are unfeasible in most circumstances. \textit{Lower Fidelity Estimation} is an Evaluation Strategy that was introduced to minimize the amount of GPU time required by each candidate through partial-training, whilst still providing a sufficiently good (albeit less robust) measure of model performance.

To prioritize exploration and leverage ABC's stochastic capabilities, HiveNAS uses LFE Evaluation Strategy. By conducting a \textit{shallow search} over the Search Space, whereby each candidate undergoes a partial training of $\epsilon < 10$ epochs, exploration is maximized. Subsequently, the best performing partially-trained candidate is then fully-trained until sufficient convergence.

\subsubsection{Framework Design, Configuration, and Reproduction} \label{subsubsec:framework_design}


The design choices for HiveNAS favor flexibility and real-world usage. By basing the framework on the expansive multi-branch Layer-based architecture Search Spaces, HiveNAS can accommodate more tasks/datasets. Furthermore, the LFE Evaluation Strategy boosts the exploration capabilities of the large and granular Search Space with minimal compromises.

Although each NAS component has a distinct role, some operations (sampling and evaluating candidates, for instance) overlap due to shared implementation properties. HiveNAS was designed with a modular and reusable pattern to encourage further research; from the dataset to the evaluation strategy, every component in the pipeline can be trivially substituted if they support particular endpoints (expanded upon in the framework's documentation\footnote{https://hivenas.readthedocs.io/en/latest/}).


Although HiveNAS was developed strictly as a NAS framework, a few numerical benchmarks are provided to test the integrity of ABC (independent of the NAS components) and to empirically infer the optimal configuration parameters for the optimization algorithm. Some alternative experiment setups are briefly discussed in Section \ref{sec:results}.

HiveNAS is implemented around a set of easily-configurable and readable operational parameters to not only be reproducible, but to also encourage further experimentation using different configurations. The generalized design pattern used in the implementation facilitates and takes into account any image dataset, where even the input size and number of classes are inferred. Additionally, the framework is open-sourced to assist with more substantial changes to the code-base and the implementation of HiveNAS variants. 


The next section showcases the results for a few alternative HiveNAS configurations. All results obtained are disclosed on the framework's public repository along with the configuration files, which can be used to reproduce the results. To further ease the generation of additional HiveNAS models, a Google Colab version of the framework is open-sourced and contains a user-friendly parameters' form that requires bare-minimum development knowledge to use.

\begin{table*}[!htb]
    \centering
    
    \small
    \begin{tabular}{M{2.0cm} M{1.75cm} M{2cm} M{1cm} M{1.75cm} M{1cm} M{2.0cm}}
        \toprule
        Model & Search Algorithm & Evaluation Strategy & GPU Usage \scriptsize(days) & CIFAR10 Error (\%) & MNIST Error (\%) & FashionMNIST Error (\%)\\
        \midrule
        
        DeepSwarm (\cite{byla_2019}) & SI & LFE & 1.25 & 11.31 & 0.39 & 6.44\\
        
        OpenNAS-ACO8 (\cite{lankford_2021}) & SI & LFE & - & 15.20 & - & 6.60\\
        
        OpenNAS-PSO10 (\cite{lankford_2021}) & SI & LFE & - & 10.00 & - & 5.70\\
        
        \midrule
        
        HiveNAS-A & & & & 11.70 & 0.43 & 6.77 \\
        
        HiveNAS-B & & & & 12.64 & - & - \\
        
        HiveNAS-C & \multirow{-3}{*}{SI} & \multirow{-3}{*}{LFE} & \multirow{-3}{*}{0.3} & \textbf{8.90} & \textbf{0.35} & \textbf{5.65}  \\
        
        \bottomrule
    \end{tabular}
    \bigskip
    \caption{Results for Swarm Intelligence-based NAS Frameworks}
    \label{tab:cifar_res_si}
\end{table*}

\begin{figure*}[!htbp]
\centering
\includegraphics[width=\textwidth]{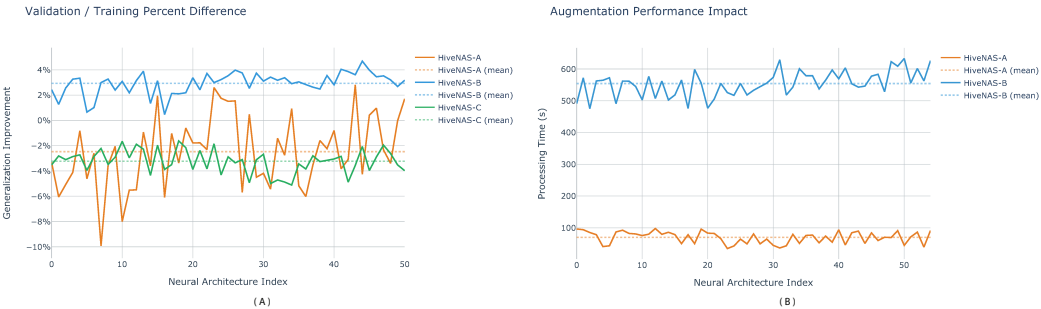}
\caption{Affine Transformations and Augmentations' performance trade-off}
\label{fig:aug_combined}
\end{figure*}

\section{Results} \label{sec:results}

In line with benchmarking conventions in the literature, HiveNAS was tested extensively on the CIFAR10 dataset as well as the MNIST and FashionMNIST datasets (to showcase its adaptive task-agnostic capabilities). With minuscule computation resources (relative to the state-of-the-art), HiveNAS shows promising potential.

In this section, 3 configurations are explored: HiveNAS-A (simple layer-based space), HiveNAS-B (simple layer-based space + pre-processing augmentations/transformations), and HiveNAS-C (compound layer-based space + residual blocks). Although Figure \ref{fig:aug_combined}(A) shows notable estimation improvement on a per-candidate basis, the impact caused by the resource-hogging augmentation/transformation pre-processing techniques impeded the overall outcome (as highlighted by Figure \ref{fig:aug_combined}(B) and Table \ref{tab:cifar_res_si}). Hence, the pre-processing step was skipped for the compound-layers variation of HiveNAS (the implementation is not deprecated from the framework, however, as it might be vital for other datasets).

All results in Table \ref{tab:cifar_res_si} have been experimented on the CIFAR10 dataset and tested using Google Colab, which runs a single GPU for a maximum of ~8 hours without interruption (a single A100 GPU). This experimentation setup, compared to the original Google Brain’s NAS setup that used 800 GPUs in parallel for 28 days (500,000 GPU hours in total), shows that HiveNAS not only yields comparable results to the current state-of-the-art, but also highlights the real-world accessibility of the framework. 

Although the MNIST dataset is often considered a trivial task for most neural networks and does not necessarily provide a robust measure of NAS performance, HiveNAS was briefly tested on it to prove that, in spite of the different input sizes/channels, the framework still adapts to the given data without manual intervention. The top performing candidate on MNIST scored \textbf{99.65\%} accuracy ($1.65$m parameters) and \textbf{94.35\%} ($0.80$m parameters) with the same configurations for FashionMNIST, both outperforming the current best SI-based NAS.



\renewcommand{\arraystretch}{1.35}
\begin{table*}[!htb]
    \centering
    
    \small
    \begin{tabular}{M{3cm} M{1.75cm} M{3cm} M{2cm} M{1.5cm} M{1.75cm}}
        \toprule
        Model & Search Algorithm & Evaluation Strategy & GPU Usage \scriptsize(days) & Params \scriptsize(millions) \\
        \midrule
        
        NAS \cite{zoph_2017} & RL & Full Training & 22,400 & 37.40 \\
        
        NASNet \cite{zoph_2018} & RL & LFE & 2,000 & 3.30 \\
        
        MetaQNN \cite{baker_2016} & RL & Full Training & 90 $\pm$ 10 & 11.20 \\
        
        PNAS \cite{liu_pnas_2018} & SMBO & LFE & - & 3.20 \\
        
        
        EAS \cite{cai_2018} & RL & Weight Inheritance & 17.5 $\pm$ 7.5 & 23.40 \\
        
        LEMONADE \cite{elsken_2018} & EA & Weight Inheritance & 40 $\pm$ 16 & 3.40 \\
        
        Efficient NAS \cite{pham_2018} & RL & OS & 0.45 & 4.60 \\
        
        \midrule
        
        HiveNAS-A &  & & & 5.30 \\
        
        HiveNAS-B &  & & & 6.62 \\
        
        HiveNAS-C & \multirow{-3}{*}{MO} & \multirow{-3}{*}{LFE} & \multirow{-3}{*}{0.3} & 1.39 \\
        
        \bottomrule
    \end{tabular}
    \bigskip
    \caption{HiveNAS computational performance on CIFAR10 relative to the state-of-the-art}
    \label{tab:rep_hivenas_performance}
\end{table*}

\begin{figure}[!htb]
\centering
\includegraphics[width=0.67\textwidth]{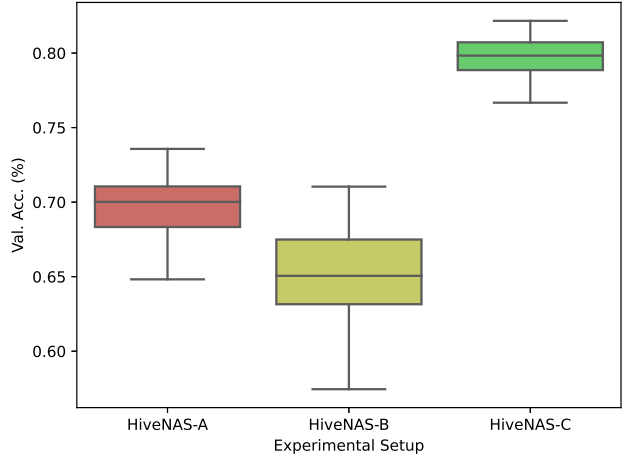}
\caption{HiveNAS candidates' validation accuracies for the initial shallow (7-epoch) search}
\label{fig:accuracies}
\end{figure}

While the HiveNAS-B setup, which uses Affine Transformations and augmentations can outperform the other setups on a per-candidate basis, image-manipulation techniques generally tend to have a significant performance footprint.

This impact on time has a consequential effect on the overall exploration capabilities of the framework; with a fixed 8-hour window of uninterrupted GPU time, much of the allowed computational time is spent on augmentation rather than evaluating other candidates and converging on optimal solutions. 

Not only does HiveNAS yields comparable results to the state-of-the-art, Table \ref{tab:rep_hivenas_performance} also shows that it does so in a fraction of the time and memory requirement, making it one of the few potential candidates for computationally-affordable NAS applications.





\section{Conclusion}


In this paper, we presented our novel NAS framework, HiveNAS, and demonstrated that SI can be used to effectively tackle NAS problems. After evaluating HiveNAS, we discovered that when compared to other similar methods, it can show competitive performance. Furthermore, we open source HiveNAS for further
development.

\subsection{Limitations \& Future Directions}

During the development phase of HiveNAS, several gaps where identified and could offer significant improvements to HiveNAS as well as other NAS frameworks.

\begin{enumerate}
    \item \textit{Robust Neighbor-Sampling}: The general consensus in the literature over the definition of ``neighbor" in discrete-space NAS is a 1-operation difference between architectures \cite{white_2021}, which does not reflect an accurate mapping of position-to-fitness. In other words, two neural topologies with 1-layer difference are most likely \textit{not} in close proximity on the fitness optimization surface, therefore the currently used definition of ``neighbors" for all Swarm Intelligence-based NAS frameworks is fundamentally erroneous and does not exploit the convergence capabilities of the optimization algorithms.
    \item \textit{Dynamic Search Space Design}: The impact the Search Space has on the overall performance of a NAS framework (as shown in Table \ref{tab:cifar_res_si} and the study conducted in \cite{yu_2020}) is substantial. Since eliminating the human factor from the ANN design process is NAS' primary goal, the Search Space design should not rely on manual input and should rather evolve to fit the given task's context.
    \item \textit{RNN Space for HiveNAS}: The current Search Space is limited to Convolutional Neural Networks, whereas ideally the type of ANN should be configurable or inferred autonomously.
\end{enumerate}

Our results show the effectiveness of ABC and its parallelization capabilities in the NAS context. Furthermore, ABC can efficiently explore the search space whilst maintaining a balanced and reasonable computational time. Our method achieves comparable results to the state-of-the-art with minimal resources, as well as being highly-configurable and open-sourced, democratizing NAS for personal usage.

\bibliography{main}


\end{document}